\documentclass[onecolumn,11pt]{IEEEtran}

\usepackage{times}
\usepackage{graphicx}
\usepackage{amsmath,amsthm}
\usepackage{amssymb,empheq}
\usepackage{latexsym}
\usepackage{epstopdf}
\usepackage{url}
\usepackage{multirow}

\usepackage{sverb}

\usepackage[ruled,linesnumbered]{algorithm2e}
\usepackage{algorithmic}

\usepackage[usenames]{color}
\definecolor{shadecolour}{gray}{0.4}

\graphicspath{{./}{./Fig/}{./fig/}{./Figs/}{./figs/}}

\newcommand{\BD}{{\mathbf{D}}}

\newcommand{\BW}{{\mathbf{W}}}

\newcommand{\BF}{{\mathbf{F}}}
\newcommand{\Bd}{{\mathbf{d}}}

\newcommand{\Bf}{{\mathbf{f}}}

\newcommand{\T}{{\!\top}}
\newcommand{\Bx}{{\mathbf{x}}}

\renewcommand{\Lambda}{\varLambda}

\usepackage{eucal}

\begin{document}


\title{Face Identification with Second-Order Pooling}

\author{
          Fumin Shen,
         Chunhua Shen 
         and
         Heng Tao Shen
\thanks
{
    F. Shen is with School of Computer Science and Engineering, University of Electronic Science and Technology of China, Chengdu 611731, P.R. China 
    (e-mail: fumin.shen@gmail.com; shenhengtao@hotmail.com).
   Part of this work was done
   when the first authour was visiting The University of Adelaide. 
  Correspondence should be addressed to F. Shen.
   }
   
   \thanks
{
    C. Shen is with The Australian Center for Visual Technologies, and School
    of Computer Science at The University of Adelaide, SA 5005, Australia (e-mail: chunhua.shen@adelaide.edu.au). 
    }
    
     \thanks
    {
    H. T. Shen is with School of Information Technology and Electrical Engineering, The University of Queensland, Australia and School of Computer Science and Engineering, University of Electronic Science and Technology of China (E-mail: shenht@itee.uq.edu.au).
    }

}

\maketitle

\begin{abstract}
%

 Automatic face recognition has received significant performance improvement
    by developing specialised facial image representations.
    On the other hand, generic object recognition has rarely been
    applied to the face recognition.
    Spatial pyramid pooling of features encoded by an over-complete dictionary has
    been the key component of many state-of-the-art image classification systems.
    Inspired by its success, in this work we develop a new face image representation
    method inspired by the second-order pooling in \cite{carreira2012semantic}, which was
    originally proposed for image segmentation.

    The proposed method
    differs from the previous methods in that,
    we encode the densely extracted local patches
%
%
    by a small-size dictionary; and the facial image signatures are obtained by pooling
    the second-order statistics of the encoded features.
    We show the importance of pooling on encoded features,
    which is bypassed by the original second-order pooling method to avoid the
    high computational cost.
    Equipped with a simple linear classifier, the proposed method
    outperforms the state-of-the-art face identification performance by large margins.
    For example, on the LFW databases, the proposed method performs better than
    the previous best by around  13\% accuracy.  

\end{abstract}

\section{Introduction}

Face identification aims to find the subject in the gallery most
similar to the probe face image. Despite decades of research effort,
        it is still an active topic in
computer vision due to both its wide applications and technical challenges. The challenges
are typically caused by various intra-class
variations (e.g., face expressions, poses, ages, image contaminations,
etc.), or lack of sufficient training data
\cite{li2011handbook}. One of the key problems is to generate a robust and discriminant
representation for facial images. Extensive research effort in the literature has been
devoted to projecting the face vectors to a low-dimensional subspace, e.g., as in the
method of eigenfaces \cite{Turk91}, Fisher-faces \cite{FisherLDA}, Laplacian faces
\cite{Laplacianfaces05}, etc. However, these holistic feature based methods often
are incapable to cope with the aforementioned problems well.

Recently, sparse representation based face classification has achieved promising results
\cite{Wright09,yang2011fisher}. Different from  previous methods,
    these methods compute the representation of the probe image  to achieve the minimum representation
error  in terms of a set of training samples or a dictionary learned from
training images.
Many algorithms have been developed in this category,
which achieve state-of-the-art performance on face recognition with image corruptions
\cite{Wright09}, face disguises \cite{yang2011fisher,RSC11} and small-size training
data \cite{ESRC2012,dengdefense2013}.

To improve the face recognition performance, many local feature based methods has been
proposed, which tend to show superior results over those based holistic features. Typical
methods in this group include
histograms of local binary patterns
(LBP) \cite{FR_LBP06}, histograms of various Gabor features \cite{GaborWavelet93,zhang2007histogram,zou2007comparative} and their fusions \cite{tan2007fusing}.
 These local feature based methods have been proven to be more robust to mis-alignment and
occlusions.

On the other hand, the local feature based image representation---bag-of-visual-words (BOV)---has
been shown state-of-the-art recognition accuracy \cite{csurka2004visual}.
The typical pipeline of BOV is: low-level local feature extraction (raw pixels, SIFT etc.),
    feature quantization
    or encoding against a pre-trained  dictionary, and descriptor
generation by spatially pooling the encoded local features. This pipeline has been
shown to achieve the state-of-the-art performance in generic image classification
\cite{grauman2005pyramid,SPM2006,yang2009linear}. Despite the success of the BOV model in
image classification, it has been rarely applied to face recognition.

The dimensionality of the learned image descriptor through the BOV pipeline is mainly determined by the size of trained dictionary (dimension of the encoded local features) and the pooling pyramid grids. It has been shown that a large dictionary size is critical to achieve a high accuracy for generic image classification \cite{coates2011analysis}.
In the meantime,
pooling features over a spatial region leads to more compact representations, and also helps to make the representation invariant to image transformation and more robust \cite{boureau2010theoretical}.
The spatial pyramid pooling model \cite{SPM2006}  has made a remarkable success, for example, in conjunction with sparse coding techniques \cite{yang2009linear}.

Average pooling and max-pooling are the two most popular pooling methods.
The latter method
usually leads to superior performance to the former one \cite{yang2009linear,boureau2010learning}.
Most previous  methods compute fist-order statistics in the pooling stage. In contrast, recently the average and max-pooling methods that incorporate the second-order
information of local features have been proposed in \cite{carreira2012semantic} for image segmentation.
Without an encoding stage, the second-order pooling strategy of \cite{carreira2012semantic}
is directly applied to the raw SIFT descriptors.
%
%

         Inspired by both the BOV model and the second-order pooling method of \cite{carreira2012semantic},
         here we propose a new method for facial image representation.
         First, local raw patches are densely extracted from the face images.
         The local patches are then encoded by an {\em small-size} dictionary, e.g., trained  by K-means.
         The encoded features are finally pooled by employing the second-order statistics over a multi-level pyramid.
         The efficacy of the learned facial features are verified by the state-of-the-art
         performance on several public benchmark databases.


    %
    %

    %
    %
The BOV model has been less frequently  studied on face recognition problems. The Fisher Vectors on densely sampled SIFT features were adopted in \cite{simonyan2013fisher} for face verification problems. In contrast, we focus on raw intensity features. Also note that no pooling is applied in this method.
Combination of sparse coding and spatial max-pooling has been used in \cite{yang2010supervised} for face recognition. However, only the first-order statistics are computed in the pooling stage. 

Our contributions mainly include:
\begin{itemize}

\item[1.]

We propose a new facial representation method based on a combination of the
BOV model and second-order pooling.
To  our knowledge, this is the first face feature extraction method using the second-order pooling technique.

\item[2.]

Different from the standard BOV methods which usually involve an over-complete dictionary, we show that, a very small number of dictionary basis are sufficient for face identification problems, in conjunction with the second-order pooling.
In contrast to the method in \cite{carreira2012semantic}, which does not apply encoding,
   we show that feature encoding is critically important
   for face identification and always improves the recognition accuracy.

\item[3.]

Coupling with a simple linear classifier, the proposed method outperforms those state-of-the-art by large margins on several benchmark databases, including AR, FERET and LFW.
In particular, the proposed method achieves perfect recognitions (100\% accuracies) on the `Fb' subset of the FERET dataset and   the `sunglasses' and `scarves' subsets of the AR dataset.
Our method obtains a higher accuracy than the best previous result by around 13\% on LFW.

\end{itemize}

\section{The proposed method}

In this section we present the details of the proposed method. We focus on extracting a discriminant representation of an image  based on its raw intensity feature other than other specific designed ones like SIFT.

\subsection{Dense local patch extraction}
\label{SEC:local}

Without loss of generality, suppose that a facial image is of $d \times d$ pixels.
As the first step,
we extract overlapped local patches of size $r \times r$  pixels with a step of $s$ pixels.
Set $l = \lfloor \frac{d - r}{s} + 1 \rfloor$,
then each image is divided into $l \times l$ patches.
Let each local patch be a
row vector $\Bx$.

It has been
shown that dense feature extraction and the pre-processing step are critical for achieving
better performance \cite{coates2011analysis}.
In practice, we extract local patches of $6 \times 6$ pixels with a stride of $1$
pixel.
 We then perform normalization on $\Bx$ as:
$
\hat{x}_i = {(x_i - m)}/{v}, 
$
where $x_i$ is the $i^{\rm th}$ element of $x$, and $m$ and $v$ are the mean and standard deviation of elements of $\Bx$.
This operation contributes to local brightness and contrast normalization as in
\cite{coates2011analysis}.

\subsection{Unsupervised dictionary training}
The goal of dictionary training is to generate a set of representative basis 
$\BD =  \{\Bd_1,\Bd_i, \cdots, \Bd_m\}  \in
\mathbf{R}^{d \times m}$. Here  $m$ is the number of atoms, and $ d $ is the input dimension.
A great deal of unsupervised dictionary learning methods has been developed, for example the  K-means clustering, sparse coding, K-SVD \cite{KSVD2006}. Dictionary can also be trained with the help of category information, such as the supervised sparse coding method \cite{yang2010supervised}.
We adopt the K-means algorithm, since it is simple and effective.
The dictionary size is a more important factor compared to the dictionary training algorithm. 

With a first-order pooling method, it has been shown that the image classification accuracy is consistently improved
as the dictionary size increases \cite{coates2011analysis}. However, this is not necessarily true when a second-order pooling technique is applied. Perhaps surprisingly, with second-order pooling, a very small number of dictionary basis are sufficient  to obtain high recognition accuracies.
We will analysis this in the  section~\ref{SEC:eval}.

As a common pre-processing step in deep learning methods, whitening  has been shown to  yield sharply localized filters when dictionary are trained by clustering on raw data \cite{coates2011analysis}. We apply the ZCA whitening on each patch \cite{ICA2000} before the dictionary learning algorithm are applied.

\subsection{Feature encoding}
With the leaned dictionary, the pre-processed local patches are are then fed into
the feature encoder to generate a set of mid-level features. Popular choices of encoding algorithms include the sparse coding \cite{yang2009linear}, Locality-constrained linear coding (LLC \cite{LLC2010}), etc. A good evaluation of different encoders can be found in \cite{Chatfield11}. The combinations of different dictionary learning and feature encoding methods are thoroughly studied in \cite{coates2011importance}. In our method, we adopt the soft threshold method, which encodes the patches by a simple feed-forward non-linearity with a fixed threshold.  This simple  encoder writes \cite{coates2011importance}:
\begin{align}
\Bf_j(\Bx) =& \max\big\{0, \BD_j^\T\Bx - \alpha \big\},\\
\Bf_{j+m}(\Bx) = &\max\big\{0, -\BD_j^\T\Bx - \alpha \big\}.
\label{EQ:ST}
\end{align}
Here $\Bf_j$ is the $j^{\mathrm{th}}$ entry of the encoded feature vector $\Bf$.
Despite its simplicity, soft threshold achieves close performance with sparse coding on the image classification task.

\subsection{Second-order pooling}
As discussed before, feature pooling plays an important role in the BOV pipeline. The pooling procedure reduces the dimensionality of the learned mid-level features on each spatial region. Moreover, the pooled features are more robust to pose variations of face images.

Depart from most of the previous methods using the first-order pooling, following \cite{carreira2012semantic}, we compute the second-order statistics of the encode features in the pooling stage. The second-order average-pooling over a spatial region $R$ is defined as:
\begin{align}
\BF_{avg} = \sum_{i:\Bf_i \in R} \Bf_i \cdot \Bf_i^{\T} / |R|,
\end{align}
where $\Bf_i$ is the column feature vector learned from region $R$ and $|R|$ is the total number of feature vectors in region $R$.  Through the outer product operation, information between all interacting pairs of descriptor dimensions is preserved. The computed $\BF_{avg}$ is a symmetric positive definite (SPD) matrix, which naturally forms a Riemannian manifold \cite{carreira2012semantic}. 
Mapping the second-order average pooling outputs by the Log-Euclidean metrics into the tangent space
 has been shown to significantly improve the classification. 
  The power normalization \cite{perronnin2010improving} is also applied after the  mapping  in \cite{carreira2012semantic}.  However, in our  experiments, this operation do not shown any  performance improvements for face identification. In practice, we only conduct the Log-Euclidean mapping:
\begin{equation}
\BF_{avg}^{log} = log(\BF_{avg}).
\end{equation}
The is computed by the algorithm in \cite{davies2003schur}.

By concatenating together all the pooled second-order statistics over a multiple-level pyramid in one vector, we obtain the final representation of a face image.
In this work, we feed the extracted features to a linear classifier to recognize the probe face image.
A simple ridge regression based multi-class classifier was used in \cite{gong2011comparing}.
We use the same linear classifier for its computational efficiency.
The ridge regression classifier has a closed-form solution, which makes the training
even faster  than the
specialized linear support vector machine (SVM) solver
{\sc Liblinear}  \cite{liblinear08}.
Despite its simplicity, the classification performance of this ridge regression approach
is on par with linear SVM \cite{gong2011comparing}.
Another benefit of this classifier is that, compared to the one-versus-all SVM, it only needs
to compute the classification matrix $\BW$ once.

\section{Do we need a large dictionary?}
\label{SEC:eval}
The key factors that affect the classification performance include the dictionary size and pooling pyramid levels. 
In addition, these two factors also determine the computational cost and the dimensionality of the learned image descriptor. In particular, the time complexity of the second-order pooling step is $\mathit{O}(m^2)$ with respect to the dictionary size $m$ (the dimensionality of the encoded local feature).

In this section, we thoroughly evaluate the impact of the components of the proposed algorithms. We vary the dictionary size from 5 to 100, as shown in Table~\ref{Tab:eval}.
Different pooling pyramids are tested: from the 3-level pyramid \{1, 2, 4\}
to a
maximum 8-level pyramid $\{1, 2, 4, 6, 8, 10, 12, 15\}$.
With this 8-level pyramid, pooling is performed on regular grids of $ 1 \times 1$, $ 2 \times 2$,
     \dots, $ 15 \times 15$ and the obtained pooled features are concatenated altogether.
The evaluation is conducted on the LFW-a dataset \cite{LFW_a}, and all the images
are down-sampled to $64 \times 64$ pixels. The dataset's description can be found in Section
\ref{SEC:exp}. We set the number of training and testing samples per subject to 5 and 2,
respectively. All the results reported in this section are based on 5 independent data
splits.

\setlength{\tabcolsep}{2pt}
\begin{table}
\centering
\caption{Cross evaluation accuracies (\%) with varying  pooling pyramids  versus dictionary sizes. All the reported results  in this table  are based on 5 independent data
splits. Here we vary the depth of pyramids from 3-level \{1, 2, 4\}
to 
maximum 8-level \{1, 2, 4, 6, 8, 10, 12, 15\}.}
\begin{tabular}{cccccccc}
\hline
pyramid&\multicolumn{7}{c}{dictionary size}\\
 levels& 5 & 10 & 20 & 40 & 60 & 80 & 100\\
 \hline

3 & $31.8  \pm  4.1$ & $76.5  \pm  1.4$ & $84.9  \pm  1.3$ & $86.2  \pm  1.8$ & $86.0  \pm  1.6$ &  $85.2  \pm  1.2$ & $84.8  \pm 1.5$ \\
4 & $66.1  \pm  3.3$ & $84,7  \pm  1.7$ & $88.0  \pm  1.4$ & $87.7  \pm  2.6$ & $86.7  \pm  2.0$ &  $ 86.5 \pm  1.9$& $86.1  \pm  1.9$\\
5 & $75.0  \pm  2.0$ & $86.5  \pm  1.6$ &  ${\bf 88.3  \pm  1.3}$ & $88.2  \pm  1.6$ & $87.7  \pm  1.3$ &  $ 88.0 \pm  1.3$& $87.7  \pm  1.1$\\
6 & $77.0  \pm  2.8$ & $86.3  \pm  1.7$ & ${\bf88.3  \pm  1.1}$ & $87.7  \pm  1.4$ & - &  -& -\\
8 & $76.5  \pm  2.7$ & $84.3  \pm  1.8$ & $86.1  \pm  1.8$ & $85.9  \pm  1.7$ &  - &  -& -\\
\hline
\end{tabular}
\label{Tab:eval}
\end{table}


As we can see from each row of Table~\ref{Tab:eval},  the best recognition result is achieved with only a small number of dictionary basis for each pyramid. For example, the proposed algorithm reach its highest average recognition rate with a dictionary of only 20  atoms and a pyramid with 5 or 6 levels. This phenomenon is very different from the literature of generic classification, where the accuracy tends to be consistently improved as the dictionary size increases and  a large over-complete dictionary  is critical to achieve high accuracies \cite{coates2011analysis}. A possible reason is that with second-order pooling, corresponding to all possible pairs of dictionary basis, the information between interacting pairs of local feature dimensions is preserved. This reduces the necessity of the use of redundant dictionary items.

From each column of Table~\ref{Tab:eval}, we can observe that more pyramid levels (up to 6) tend to result in better performance. The information of the pooled representation is enriched from multi-scale patches. However redundant features do not always improve the final recognition rates.  Accuracy drops are also observed with more than 8 pyramid levels, similar as the dictionary size. 

Taken into account both the performance and computation cost, we set in the following paper the dictionary size and number of pyramid levels to 20 and 5, respectively.

\section{Pooling on encoded features or raw patches?}
In the previous BOV methods using the first-order pooling, the encoding stage is  usually in conjunction with an over-complete dictionary, which consequently produces high-dimensional local features.
Different from the first-order pooling methods, second-order pooling results in much larger computational complexities due to the outer product operations. To avoid this, the method in \cite{carreira2012semantic} choose to directly apply pooling on the raw local descriptors (e.g., SIFT) without any encoding stage. The authors claim that good performance can be obtained without any feature coding due to that the preservation of the second-order statistics.

In this section, we will explore
whether feature encoding is critical for face identification, based on raw local patches. Table~\ref{Tab:encoding} shows the identification rates of the second-order method with and without feature encoding on LFW and FERET. The number of training and testing samples are set to 5 and 2 respectively on both of the two databases. The datasets' description can be found in Section
\ref{SEC:exp}.

\setlength{\tabcolsep}{3pt}
\begin{table}
\centering
\caption{Identification accuracy (\%) of the second-order method with and without feature encoding on LFW and FERET. A 5-level pyramid is used.}
\begin{tabular}{c|cc}
\hline
 & with encoding & without encoding\\
 \hline \hline
 LFW & $88.3 \pm 1.3$ & $82.8 \pm 2.4$\\
 FERET & $98.5 \pm 0.3$ & $95.0 \pm 1.0$\\
 \hline
\end{tabular}
\label{Tab:encoding}
\end{table}

 It is clear that the classification accuracy is largely improved by second-order pooling  encoded features than pooling  raw features.
In this case, the encoded features provide more discriminant information than the raw patches. Therefore, for face identification problems we suggest using the second-order pooling method on the encoded local features, generated from a dictionary which is not necessarily large.

\section{Experimental results}
\label{SEC:exp}

In this section, we thoroughly evaluate the proposed method on several public facial image datasets including FERET \cite{FERET},  AR \cite{AMM98}, LFW \cite{LFWTech} and Pubfig83 \cite{pubfig83}.
Since there are a great deal of algorithms developed in the face identification community, we only compare our method with a few of them, which are representing or reported to achieve the state-of-the-art results. These methods include the  superposed sparse representation classifier (SSRC \cite{dengdefense2013}),  local patch based Volterra \cite{kumar2012trainable}, multi-scale patch based MSPCRC
\cite{zhu2012multi}, and the popular BOV model Locality-constrained linear coding (LLC \cite{LLC2010}). The soft thresholding (ST) method \cite{coates2011importance} with the first-order pooling is also taken into comparison. 
Both LLC and soft thresholding use a 3-level pyramid and 1024 dictionary size. 
We set the regularization parameter $\lambda$ to 0.001 for SRC, RSC, MSPCRC and LLC, and 0.005 for SSRC, according to the authors' recommendation. The parameter $\alpha$ is set to 0.25 for ST. For fair comparisons, contrast normalization and ZCA whitening are performed for both LLC, ST and our method.

\subsection{FERET}
The FERET dataset \cite{FERET} is a widely-used standard face recognition benchmark set provided by DARPA. Since the images in FERET are collected in multiple sessions during several years, they have complex intraclass variability.
In our first experiment on FERET, we apply the standard protocols. With the gallery Fa (1196 images of 1196 subjects), we test on four probe sets: Fb (1195 images of 1195 subjects), Fc (194 images of 194 subjects), Duplicate I (722 images of 243 subjects, denoted as DupI), and Duplicated II (234 images of 75 subjects, denoted as DupII). Fb and Fc are captured with expression and illumination variations respectively. DupI and DupII are captured at different times. All the images are aligned based on the manually located eye centers and normalized to $150 \times 130$ pixels. 

In this experiment, we mainly compare our facial image representation method to other feature extraction algorithms. The first five methods in Table~\ref{Tab:FERET} are specifically designed for face feature extraction, which are based on Gabor feature \cite{zhang2007histogram,zou2007comparative} or feature fusion with different features feature \cite{tan2007fusing,yang2012monogenic,xie2010fusing}.
LLC and ST are two BOV methods.

The compared results are shown in Table~\ref{Tab:FERET}.  It is clear that, on all four sessions the proposed method achieves the best performance. Among the five specific facial representation methods, MBC   obtains superior results on Fb and Fc, while  Xie's method performs better on DupI and DupII. However, they are still inferior to our method, which is much simpler.
We can  also see that all the three BOV approaches LLC, ST and our method perform very well on the subset Fb and Fc with only expression and illumination variations. However, on the subset DupI and DupII with images taken at different years, LLC and ST obtain dramatic performance drop, while our second-order method still get very high accuracies. This demonstrates the discriminant ability of the proposed face image representation method.

\begin{table}
\centering
\caption{Recognition accuracies (\%) on FERET with standard protocols. Results of the first 5 methods are cited from \cite{yang2012monogenic} with identical settings. }
\begin{tabular}{c|cccc}
\hline 
Session & Fb(expression) & Fc (illumination)& DupI (aging) & DupII (aging)\\
\hline\hline
HGPP \cite{zhang2007histogram} & 97.5 & 99.5 & 79.5 & 77.8\\
Zou's method \cite{zou2007comparative} & 99.5 & 99.5 & 85.0 & 79.5\\
Tan's method \cite{tan2007fusing}& 98.0 & 98.0 & 90.0 & 85.0\\
Xie's method \cite{xie2010fusing} & 99.0 & 99.0 & 94.0 & 93.0\\
MBC \cite{yang2012monogenic} & 99.7 & 99.5 & 93.6 & 91.5\\
\hline
LLC & 99.6 & \textbf{100} & 78.4 & 69.2\\
ST &98.8 &99.5 &70.5 &68.0 \\
Ours & \textbf{99.8}&\textbf{100}&\textbf{96.0} & \textbf{96.6}\\
\hline
\end{tabular}
\label{Tab:FERET}
\end{table}

 In the second experiment on FERET, we used a subset of FERET which includes images with pose variations from 200 subjects. Each individual contains 7 samples with  pose variations of up to 25 degrees. It is composed of images whose names are marked with `ba', `bj', `bk', `be', `bf', `bd' and `bg'. These images are cropped and resized to $80 \times 80$ pixels \cite{yang2003combined}. We randomly select 5 samples for training and 2 samples for testing. The mean results of 5 independent runs with images down-sampled to $64 \times 64$ pixels are shown in Table~\ref{Tab:FERET-pose}.
 
We can clearly see that our method achieves the highest recognition rate. In particular, our method obtains an accuracy of 98.5\% which is  higher than the second best SSRC by 16.4\%. RSC dose not show better results than SRC. The local patch based MSPCRC and Volterra do not perform well. This is probably because they are incapable of coping with pose variations.

\begin{table}
\centering
\caption{Recognition accuracies (\%) on FERET with pose variations. Results are based on 5 independent runs.}
\begin{tabular}{ccccccc}
\hline
 Method &SRC & RSC & MSPCRC & Volterra& SSRC & Ours\\ 
 \hline
 Accuracy &$73.8 \pm 2.0$ & $73.4 \pm 2.6$&$42.3 \pm 2.8$ &$50.4 \pm 2.3$&$82.1 \pm 1.6$&$\bf 98.5 \pm 0.3$\\
 \hline
\end{tabular}
 \label{Tab:FERET-pose}
\end{table}

\subsection{AR}

Since AR has been used very often to evaluate face recognition algorithms, we compare our methods to several published state-of-the-art results on this dataset.
The AR dataset has 126 subjects (70 men and 56 women) and contains more than 4000 facial images. Each subject contains 26 images taken in two separate sessions. The images exhibit a number of variations including facial expression (neutral, smile, anger, scream), illumination (left light on, right light on, both sides light on) and occlusion (sunglasses, scarves). We select 100 subjects (50 men and 50 women) in our experiment.
Four different situations are tested. For the `All' situation, all the 13 images in the first session are used for training and the other 13 images in the second session for testing. For the `clean' situation, 7 samples in each session with only illumination and expression changes are used for training and testing. For the `sunglasses' and `scarf' situation, 8 clean samples from two sessions are used for training and 2 images with sunglasses and scarf are used for testing. All the images are resized to $64 \times 64$ pixels. The test results are shown in Table~\ref{Tab:AR}.

It is obvious that our method achieves the highest accuracies in all situations. In particular, our method achieves perfect recognitions (100\%) when face images are with sunglasses and scarves occlusions on this dataset. To the best our knowledge, our method is the only one achieving this results in both occlusion situations. Again, our method show its discriminant ability for face images.

\begin{table}
\centering
\caption{Comparison with recent state-of-the-art results (\%) on AR with four different settings.
     For the first six methods, we quote the best reported results from the corresponding published papers, with the same experiment setting.}
\begin{tabular}{r | cccc}
\hline
Situation & Clean   & Sunglasses  & Scarves& All   \\
\hline\hline
SRC \cite{Wright09}&92.9&87.0&57.5&-\\
RSC \cite{RSC11} & 96.0  & 99.0 & 97.0& -\\
DRDL \cite{ma2012sparse} & 95.0 & - &- &-\\
$L_2$ \cite{Javen2011} & -  & 78.5 & 79.5& 95.9\\
SSRC \cite{dengdefense2013} & -  & 90.9 &90.9& 98.6\\
FDDL \cite{yang2011fisher} & 92.0 & - &- &-\\
Volterra &90.9&96.1&92.1&87.5\\
\hline
ST & 
98.1&
99.5&
98.5&
99.2\\
LLC&
98.8&
99.0&
99.0&
99.3\\
Ours & \textbf{99.9}&\textbf{100}&\textbf{100}&\textbf{99.9}\\
\hline
\end{tabular}
\label{Tab:AR}
\end{table}

\subsection{LFW}
Following the settings in \cite{zhu2012multi},
     here we use LFW-a, an aligned version of LFW using commercial face alignment software \cite{LFW_a}.
     A subset of LFW-a including 158 subjects with each subject more than 10 samples are used. The images are cropped to $121 \times 121$ pixels.
      We randomly selected 5 samples for training and another 2 samples for
testing. All the images are finally resized to $32 \times 32$ pixels
as  in \cite{zhu2012multi}.

Table~\ref{Tab:LFW} lists the compared identification accuracies.
Consistent with the previous results, our method outperforms all other algorithms by even larger gaps on this very challenging dataset. We obtain higher accuracies than  MSPCRC and ST by  29.7\% and 4.8\%, respectively.  Table~\ref{Tab:LFW_Dim} shows the compared results with different image sizes. If we use a larger image size $64 \times 64$, the accuracy of our method will increase to 88.3\%, while ST and MSPCRC do not shown any improvements. LLC achieves better results with larger image dimension, however it is still inferior to our method.

\setlength{\tabcolsep}{2pt}
\begin{table}
\centering
\caption{Recognition accuracy on LFW with downsampled $32 \times 32$ images.
     Results are based on 20 independent runs. The results for PNN, Volterra and MSPCRC are quoted from \cite{zhu2012multi} with identical settings.}
\begin{tabular}{cccccccc}
\hline
Method & SRC&PNN\cite{kumar2011maximizing}&Volterra&MSPCRC & ST & LLC & Ours\\\hline
Accuracy& $44.1 \pm 2.6$ & $47.4 \pm 2.7$& $40.3 \pm 2.7$& $49.0 \pm 2.9$& $73.9 \pm 1.6$& $64.1 \pm 2.2$ &${\bf 78.7 \pm 3.1}$\\
\hline
\end{tabular}
\label{Tab:LFW}
\end{table}

\begin{table}
\centering
\caption{Recognition accuracy (\%) on LFW-a with different image sizes.}
\begin{tabular}{r  cc}
\hline
Method & $32 \times 32$ pixels& $64 \times 64$ pixels\\
\hline
MSPCRC &$49.0 \pm 2.9$&$47.7 \pm 2.0$\\
LLC & $64.1 \pm 2.2$ & $75.4 \pm 2.6$ \\
ST & $73.9 \pm 1.6$& $73.7 \pm 2.0$ \\
Ours & ${\bf 78.7 \pm 3.1}$ & ${\bf  88.3 \pm 1.3}$  \\
\hline
\end{tabular}
\label{Tab:LFW_Dim}
\end{table}

\subsection{Pubfig83}

PubFig83 \cite{pubfig83} is a subset of PubFig dataset \cite{kumar2009attribute}, which include a large collection of real-world images of celebrities
collected from the Internet.
With 100 more samples in each of the 83 individual, PubFig83 is reconfigured for the problem of unconstrained face identification.  We follow the evaluation protocol of \cite{pubfig83} and use 90 samples for training, 10 samples for testing. The images are resized to $100 \times 100$ pixels and the rest results are based on 10 independent runs. 
The results are shown in Table~\ref{Tab:Pubfig83}. 

The first four results in Table~\ref{Tab:Pubfig83} are quoted from \cite{chiachia2012person} using  different subspace method and combined representations of  LBP, HOG and Gabor. RAW refers to the raw image representation, PLS refers to the partial least squares and PS-PLS refers to person specific PLS \cite{chiachia2012person}.
It is clear that our method outperforms all the other methods. In specific, the proposed method outperforms the best  of other methods  by 2.6\%. 

\setlength{\tabcolsep}{3pt}
\begin{table}
\centering
\caption{Recognition accuracy (\%) on Pubfig83. The first four methods  use  combined representations of  LBP, HOG and Gabor \cite{chiachia2012person} and linear SVMs. }
\begin{tabular}{cccccc}
\hline
  Method &  RAW & PCA & PLS & PS-PLS 
& Ours\\
Accuracy & $82.6 \pm 0.3$ &$82.4 \pm 0.3$ &$83.0 \pm 0.3$ & $85.4 \pm 0.3$  & ${\bf 88.0 \pm 0.8}$\\
\hline
\end{tabular}
\label{Tab:Pubfig83}
\end{table}

\section{Conclusion}

In this paper, we propose a new representation method for facial images, inspired by the second-order pooling method in \cite{carreira2012semantic}. The key idea is to extract the second-order statistics of the encoded local features, which are generated by a small-size dictionary. We directly apply this method on densely extracted local patches other than specialized features like SIFT. We show that, with second-order pooling, the dictionary size (20 in all our experiments) is not necessarily as large as in the first-order method.   We also show that the feature encoding procedure is critical for face identification problems, even if the dictionary is very small. 
The discriminant power of the proposed facial image representation methods has been verified by the state-of-the-art performance on several benchmark datasets. 
On the FERET database, for instance, our method achieves  accuracies of 100\%, 96.0\% and 96.6\% on the Fc, DupI and DupII subsets, respectively. The proposed method also outperforms the best report results on AR and LFW dataset on face identification problems.

We plan to explore the efficacy of the proposed image representation method on face verification problems in the future work.


\begin{thebibliography}{10}

\bibitem{carreira2012semantic}
Carreira, J., Caseiro, R., Batista, J., Sminchisescu, C.:
\newblock Semantic segmentation with second-order pooling.
\newblock In: Proc. Eur. Conf. Comp. Vis.
\newblock Springer (2012)  430--443

\bibitem{li2011handbook}
Li, S.Z., Jain, A.K.:
\newblock Handbook of Face Recognition.
\newblock Springer London (2011)

\bibitem{Turk91}
Turk, M., Pentland, A.:
\newblock Eigenfaces for recognition.
\newblock J. Cognitive Neuroscience \textbf{3}(1) (January 1991)  71--86

\bibitem{FisherLDA}
Belhumeur, P.N., Hespanha, J.a.P., Kriegman, D.J.:
\newblock Eigenfaces vs. fisherfaces: Recognition using class specific linear
  projection.
\newblock IEEE Trans. Patt. Anal. Mach. Intell. \textbf{19} (July 1997)
  711--720

\bibitem{Laplacianfaces05}
He, X., Yan, S., Hu, Y., Niyogi, P., Zhang, H.J.:
\newblock Face recognition using laplacianfaces.
\newblock IEEE Trans. Patt. Anal. Mach. Intell. \textbf{27}(3) (march 2005)
  328 --340

\bibitem{Wright09}
Wright, J., Yang, A.Y., Ganesh, A., Sastry, S.S., Ma, Y.:
\newblock Robust face recognition via sparse representation.
\newblock IEEE Trans. Pattern Anal. Mach. Intell. \textbf{31} (2009)  210--227

\bibitem{yang2011fisher}
Yang, M., Zhang, L., Feng, X., Zhang, D.:
\newblock Fisher discrimination dictionary learning for sparse representation.
\newblock In: Proc. IEEE Int. Conf. Comp. Vis. (2011)  543--550

\bibitem{RSC11}
Yang, M., Zhang, L., Yang, J., Zhang, D.:
\newblock Robust sparse coding for face recognition.
\newblock In: Proc. IEEE Conf. Comp. Vis. Patt. Recogn. (2011)  625--632

\bibitem{ESRC2012}
Deng, W., Hu, J., Guo, J.:
\newblock {Extended SRC}: undersampled face recognition via intra-class variant
  dictionary.
\newblock IEEE Trans. Pattern Anal. Mach. Intell. \textbf{34}(9) (2012)
  1864--1870

\bibitem{dengdefense2013}
Deng, W., Hu, J., Guo, J.:
\newblock In defense of sparsity based face recognition.
\newblock In: Proc. IEEE Conf. Comp. Vis. Patt. Recogn. (2013)  399--406

\bibitem{FR_LBP06}
Ahonen, T., Hadid, A., Pietikainen, M.:
\newblock Face description with local binary patterns: application to face
  recognition.
\newblock IEEE Trans. Pattern Anal. Mach. Intell. \textbf{28}(12) (2006)
  2037--2041

\bibitem{GaborWavelet93}
Lades, M., Vorbruggen, J., Buhmann, J., Lange, J., von~der Malsburg, C., Wurtz,
  R., Konen, W.:
\newblock Distortion invariant object recognition in the dynamic link
  architecture.
\newblock IEEE Trans. Comput. \textbf{42}(3) (1993)  300 --311

\bibitem{zhang2007histogram}
Zhang, B., Shan, S., Chen, X., Gao, W.:
\newblock Histogram of gabor phase patterns (hgpp): a novel object
  representation approach for face recognition.
\newblock IEEE Trans. Image Processing \textbf{16}(1) (2007)  57--68

\bibitem{zou2007comparative}
Zou, J., Ji, Q., Nagy, G.:
\newblock A comparative study of local matching approach for face recognition.
\newblock IEEE Trans. Image Processing \textbf{16}(10) (2007)  2617--2628

\bibitem{tan2007fusing}
Tan, X., Triggs, B.:
\newblock Fusing gabor and {LBP} feature sets for kernel-based face
  recognition.
\newblock In: International Workshop on Analysis and Modeling of Faces and
  Gestures. (2007)  235--249

\bibitem{csurka2004visual}
Csurka, G., Dance, C., Fan, L., Willamowski, J., Bray, C.:
\newblock Visual categorization with bags of keypoints.
\newblock In: Workshop on statistical learning in computer vision, ECCV.
  Volume~1. (2004)  1--2

\bibitem{grauman2005pyramid}
Grauman, K., Darrell, T.:
\newblock The pyramid match kernel: Discriminative classification with sets of
  image features.
\newblock In: Proc. IEEE Conf. Comp. Vis. Volume~2. (2005)  1458--1465

\bibitem{SPM2006}
Lazebnik, S., Schmid, C., Ponce, J.:
\newblock Beyond bags of features: Spatial pyramid matching for recognizing
  natural scene categories.
\newblock In: Proc. IEEE Conf. Comp. Vis. Patt. Recogn. (2006)  2169--2178

\bibitem{yang2009linear}
Yang, J., Yu, K., Gong, Y., Huang, T.:
\newblock Linear spatial pyramid matching using sparse coding for image
  classification.
\newblock In: Proc. IEEE Conf. Comp. Vis. Patt. Recogn. (2009)  1794--1801

\bibitem{coates2011analysis}
Coates, A., Ng, A., Lee, H.:
\newblock An analysis of single-layer networks in unsupervised feature
  learning.
\newblock In: Proc. Int. Conf. Artif. Intell. Stat. (2011)  215--223

\bibitem{boureau2010theoretical}
Boureau, Y.L., Ponce, J., LeCun, Y.:
\newblock A theoretical analysis of feature pooling in visual recognition.
\newblock In: Proc. Int. Conf. Mach. Learn. (2010)  111--118

\bibitem{boureau2010learning}
Boureau, Y., Bach, F., LeCun, Y., Ponce, J.:
\newblock Learning mid-level features for recognition.
\newblock In: Proc. IEEE Conf. Comp. Vis. Patt. Recogn. (2010)  2559--2566

\bibitem{simonyan2013fisher}
Simonyan, K., Parkhi, O.M., Vedaldi, A., Zisserman, A.:
\newblock Fisher vector faces in the wild.
\newblock In: Proc. British Mach. Vis. Conf. Volume~1. (2013) ~7

\bibitem{yang2010supervised}
Yang, J., Yu, K., Huang, T.:
\newblock Supervised translation-invariant sparse coding.
\newblock In: Proc. IEEE Conf. Comp. Vis. Patt. Recogn., IEEE (2010)
  3517--3524

\bibitem{KSVD2006}
Aharon, M., Elad, M., Bruckstein, A.:
\newblock {K-SVD}: an algorithm for designing overcomplete dictionaries for
  sparse representation.
\newblock IEEE Trans. Signal Processing \textbf{54}(11) (2006)  4311--4322

\bibitem{ICA2000}
Hyvarinen, A., Oja, E.:
\newblock Independent component analysis: algorithms and applications.
\newblock Neural Netw. \textbf{13}(4-5) (2000)  411--430

\bibitem{LLC2010}
Wang, J., Yang, J., Yu, K., Lv, F., Huang, T., Gong, Y.:
\newblock Locality-constrained linear coding for image classification.
\newblock In: Proc. IEEE Conf. Comp. Vis. Patt. Recogn. (2010)  3360 --3367

\bibitem{Chatfield11}
Chatfield, K., Lempitsky, V., Vedaldi, A., Zisserman, A.:
\newblock The devil is in the details: an evaluation of recent feature encoding
  methods.
\newblock In: Proc. British Mach. Vis. Conf. (2011)

\bibitem{coates2011importance}
Coates, A., Ng, A.:
\newblock The importance of encoding versus training with sparse coding and
  vector quantization.
\newblock In: Proc. Int. Conf. Mach. Learn. (2011)  921--928

\bibitem{perronnin2010improving}
Perronnin, F., S{\'a}nchez, J., Mensink, T.:
\newblock Improving the fisher kernel for large-scale image classification.
\newblock In: Proc. Eur. Conf. Comp. Vis.
\newblock Springer (2010)  143--156

\bibitem{davies2003schur}
Davies, P.I., Higham, N.J.:
\newblock A schur-parlett algorithm for computing matrix functions.
\newblock SIAM Journal on Matrix Analysis and Applications \textbf{25}(2)
  (2003)  464--485

\bibitem{gong2011comparing}
Gong, Y., Lazebnik, S.:
\newblock Comparing data-dependent and data-independent embeddings for
  classification and ranking of internet images.
\newblock In: Proc. IEEE Conf. Comp. Vis. Patt. Recogn. (2011)  2633--2640

\bibitem{liblinear08}
Fan, R., Chang, K., Hsieh, C., Wang, X., Lin, C.:
\newblock Liblinear: a library for large linear classification.
\newblock J. Mach. Learn. Res. (2008)  1871--1874

\bibitem{LFW_a}
Wolf, L., Hassner, T., Taigman, Y.:
\newblock Similarity scores based on background samples.
\newblock In: Proc. Asian conf. Comp. Vis. (2010)  88--97

\bibitem{FERET}
Phillips, P., Wechsler, H., Huang, J., Rauss, P.:
\newblock The {FERET} database and evaluation procedure for face-recognition
  algorithms.
\newblock Image Vis. Comput. \textbf{16}(5) (1998)  295--306

\bibitem{AMM98}
Martinez, A., Benavente, R.:
\newblock {The AR Face Database}.
\newblock CVC, Tech. Rep. (1998)

\bibitem{LFWTech}
Huang, G.B., Ramesh, M., Berg, T., Learned-Miller, E.:
\newblock Labeled faces in the wild: A database for studying face recognition
  in unconstrained environments.
\newblock Technical Report 07-49, University of Massachusetts, Amherst (October
  2007)

\bibitem{pubfig83}
Pinto, N., Stone, Z., Zickler, T., Cox, D.D.:
\newblock {Scaling-up Biologically-Inspired Computer Vision: A Case-Study on
  Facebook}.
\newblock In: Workshop on Biologically Consistent Vision, IEEE Conf. Comp. Vis.
  Patt. Recogn. (2011)

\bibitem{kumar2012trainable}
Kumar, R., Banerjee, A., Vemuri, B.C., Pfister, H.:
\newblock Trainable convolution filters and their application to face
  recognition.
\newblock IEEE Trans. Pattern Anal. Mach. Intell. \textbf{34}(7) (2012)
  1423--1436

\bibitem{zhu2012multi}
Zhu, P., Zhang, L., Hu, Q., Shiu, S.:
\newblock Multi-scale patch based collaborative representation for face
  recognition with margin distribution optimization.
\newblock In: Proc. Eur. Conf. Comp. Vis. (2012)  822--835

\bibitem{yang2012monogenic}
Yang, M., Zhang, L., Shiu, S.K., Zhang, D.:
\newblock Monogenic binary coding: An efficient local feature extraction
  approach to face recognition.
\newblock IEEE Trans. Trans. Inf. Forensics Security \textbf{7}(6) (2012)
  1738--1751

\bibitem{xie2010fusing}
Xie, S., Shan, S., Chen, X., Chen, J.:
\newblock Fusing local patterns of gabor magnitude and phase for face
  recognition.
\newblock IEEE Trans. Image Processing \textbf{19}(5) (2010)  1349--1361

\bibitem{yang2003combined}
Yang, J., Yang, J., Frangi, A.:
\newblock Combined {Fisherfaces} framework.
\newblock Image Vis. Comput. \textbf{21}(12) (2003)  1037--1044

\bibitem{ma2012sparse}
Ma, L., Wang, C., Xiao, B., Zhou, W.:
\newblock Sparse representation for face recognition based on discriminative
  low-rank dictionary learning.
\newblock In: Proc. IEEE Conf. Comp. Vis. Patt. Recogn. (2012)  2586--2593

\bibitem{Javen2011}
Shi, Q., Eriksson, A., van~den Hengel, A., Shen, C.:
\newblock Is face recognition really a compressive sensing problem?
\newblock In: Proc. IEEE Conf. Comp. Vis. Patt. Recogn. (2011)  553--560

\bibitem{kumar2011maximizing}
Kumar, R., Banerjee, A., Vemuri, B., Pfister, H.:
\newblock Maximizing all margins: pushing face recognition with kernel
  plurality.
\newblock In: Proc. IEEE Int. Conf. Comp. Vis. (2011)  2375--2382

\bibitem{kumar2009attribute}
Kumar, N., Berg, A.C., Belhumeur, P.N., Nayar, S.K.:
\newblock Attribute and simile classifiers for face verification.
\newblock In: Proc. IEEE Conf. Comp. Vis., IEEE (2009)  365--372

\bibitem{chiachia2012person}
Chiachia, G., Pinto, N., Schwartz, W.R., Rocha, A., Falc{\~a}o, A.X., Cox,
  D.D.:
\newblock Person-specific subspace analysis for unconstrained familiar face
  identification.
\newblock In: Proc. British Mach. Vis. Conf. (2012)  1--12

\end{thebibliography}
\end{document}